\pdfoutput=1

\documentclass[11pt]{article}

\usepackage[]{acl}

\usepackage{times}
\usepackage{latexsym}

\usepackage[T1]{fontenc}
\usepackage{multicol}
\usepackage[utf8]{inputenc}

\usepackage{microtype}
\usepackage{enumitem}
\usepackage{graphicx}
\usepackage[export]{adjustbox}
\usepackage{amsmath}

\definecolor{positive}{HTML}{466da0}
\definecolor{negative}{HTML}{B44278}

\usepackage{siunitx,tabularx,ragged2e,booktabs,caption}
\usepackage{array}
\usepackage{enumitem}
\usepackage{amsmath,amssymb}
\usepackage{colortbl}
\usepackage{multirow}
\usepackage{booktabs}
\usepackage{pifont}
\usepackage{pgfplots}
\usepackage{parskip}
\usepackage{subcaption}
\usepackage{microtype}
\newcolumntype{P}[1]{>{\raggedright\arraybackslash}p{#1}}
\title{Small But Funny: A Feedback-Driven Approach to Humor Distillation}

\usepackage{array}
\usepackage{enumitem}
\usepackage{amsmath,amssymb}
\usepackage{colortbl}
\usepackage{multirow}
\usepackage{subcaption}

\usepackage{latexsym}
\usepackage{subcaption}
\usepackage{algorithm}
\usepackage{algorithmicx}

\usetikzlibrary{calc,graphs,shapes.misc, positioning,backgrounds,shapes, arrows}

\author{Sahithya Ravi$^{1,3}$~~Patrick Huber$^{2} $~~Akshat Shrivastava$^{2} $~~\textbf{Aditya Sagar}$^{2} $~~\\\textbf{Ahmed Aly}$^{2}$~~\textbf{Vered Shwartz}$^{1,3} $~~\textbf{Arash Einolghozati}$^{2}$ \\
$^1$ University of British Columbia\\$^2$ Meta AI\\$^3$ Vector Institute for AI\\
{\tt \small\{sahiravi, vshwartz\}@cs.ubc.ca}\\
{\tt \small\{patrickHuber, akshats, adithyasagar, ahhegazy, arash\}@meta.com}
}

\newcommand{\eg}{{\em e.g.}}

\newcommand{\quotes}[1]{``#1''}
\newcommand{\textunderscript}[1]{$_{\text{#1}}$}
\begin{document}
\maketitle
\begin{abstract}
The emergence of Large Language Models (LLMs) has brought to light promising language generation capabilities, particularly in performing tasks like complex reasoning and creative writing. Consequently, distillation through imitation of teacher responses has emerged as a popular technique to transfer knowledge from LLMs to more accessible, Small Language Models (SLMs). While this works well for simpler tasks, there is a substantial performance gap on tasks requiring intricate language comprehension and creativity, such as humor generation.   
We hypothesize that this gap may stem from the fact that creative tasks might be hard to learn by imitation alone and explore whether an approach, involving supplementary guidance from the teacher could yield higher performance. 
To address this, we study the effect of assigning a dual role to the LLM -- as a ``teacher'' generating data, as well as a ``critic'' evaluating the student's performance. Our experiments on humor generation reveal that the incorporation of feedback significantly narrows the performance gap between SLMs and their larger counterparts compared to merely relying on imitation. As a result, our research highlights the potential of using feedback as an additional dimension to data when transferring complex language abilities via distillation. 

\end{abstract}

\section{Introduction}
\label{sec:intro}
NLP is on a trajectory towards creating increasingly large models \cite{openai2023gpt4, Touvron2023Llama2O}. LLMs achieve high performance across many tasks in both 
zero and few-shot settings. 
However, there are growing concerns about the computational efficiency and environmental sustainability of such approaches 
\cite{strubell-etal-2019-energy}.

Knowledge distillation \cite{Hinton2015DistillingTK} has thus gained a renewed interest, where the term has evolved to denote the process of distilling the responses of LLMs to SLMs \cite{west-etal-2022-symbolic}. Recent work has explored the distillation of commonsense knowledge \cite{bhagavatula-etal-2023-i2d2}, chain-of-thought reasoning \cite{li-etal-2023-symbolic}, and summarization abilities \cite{liu-chen-2022-learning, jung2023impossible} to smaller language models. However, the exploration of distilling creative abilities such as humor into SLMs remains an open and challenging area of research. 
\begin{figure}[t!]
  \centering
  \includegraphics[width=0.45\textwidth]{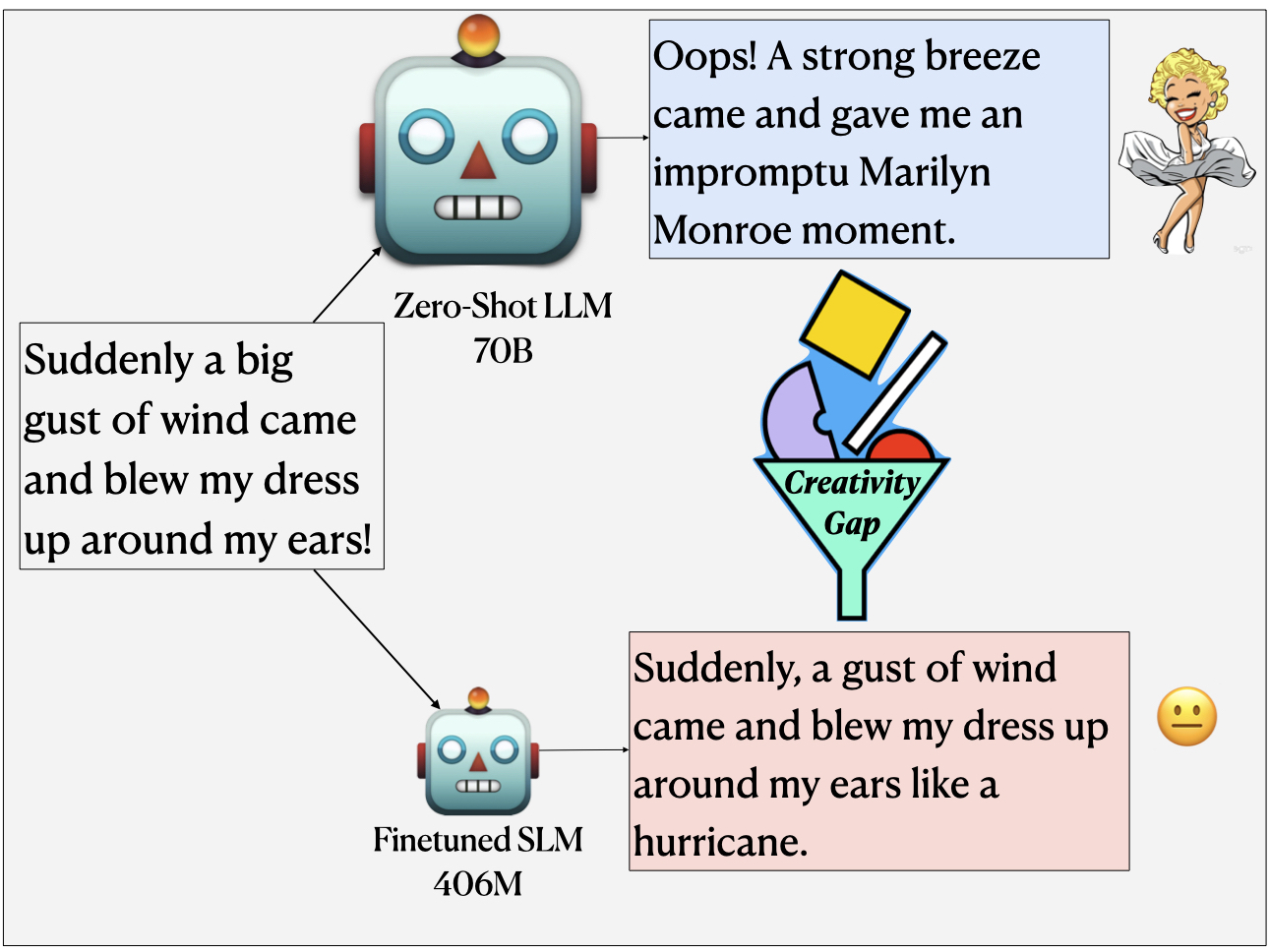}
  \caption{Performance gap between LLMs and SLMs: Generations from a teacher LLM (Llama2) and a student SLM (BART) finetuned on its outputs.} 
  \label{fig:motivation}
\end{figure}
In this work, we explore the application of knowledge distillation in the context of the creative style-transfer task of \emph{conditional humor generation}. Given a literal text, the goal is to generate a humorous meaning-preserving paraphrase, as shown in Figure~\ref{fig:motivation}. 
While LLMs do not match the subtleties of humor in human-written text \cite{hessel-etal-2023-androids, Chakrabarty2023ArtOA}, they surpass their smaller counterparts 
\cite{Radford2019LanguageMA, chen2023minigptv2, hessel-etal-2023-androids}. 

We argue that such creative tasks are challenging for SLMs to learn.  First, due to the inherent constraints of SLMs, such as reduced model capacity, they are limited in their ability to explore diverse solution spaces and generate innovative outputs.
Second, although imitating a teacher model 
is a good starting point,  it may result in superficial overfitting to the teacher's style rather than learning the task itself  
\cite{gudibande2023false}. Fig.~\ref{fig:motivation} demonstrates that even after fine-tuning on the humorous responses from the LLM, the SLM outputs fall flat. Existing techniques that address this gap attempt to improve task understanding through distilling chains of thoughts  \cite{mukherjee2023orca, wang-etal-2023-scott}. This approach is less applicable for creative generation tasks, which can't be scripted. For instance, it is possible to solve a math reasoning problem systematically, but it is difficult to come up with a recipe for a joke \cite{hessel-etal-2023-androids, west2023generative}. 

To bridge this gap, and improve upon mimicry of the teacher, we propose a novel distillation framework 
involving both imitation and feedback. Following the typical imitation stage in which the SLM learns from the LLM's outputs, 
we use the LLM as a critic to provide feedback on the student's outputs, facilitating iterative improvement.  

Evaluation of our student models on the proposed task based on EmpatheticDialogues  \cite{rashkin-etal-2019-towards} and  Samsum \cite{gliwa-etal-2019-samsum} datasets confirms the advantage of learning from feedback; our student, based on the small BART model,  performs on par with LLMs that are orders of magnitude larger, such as Llama2-70B upto 65\% of the time, and significantly outperforms supervised fine-tuning by a margin of 18-20\% .  We assess the strengths and limitations of the critic in evaluating the SLM by comparing it against human judgments. We found that our critics can match human judgments with up to 76\% accuracy, but can also suffer from biases due to length, position, or other biases. We explore the effect of data size,  frequency of critic intervention, and the effect of potential evaluation biases on narrowing the gap between the SLM and LLM. 
 
Our work on distilling humor is a step towards more natural and engaging conversations \cite{Ritchie2007APA}, making SLMs more appealing for downstream applications where latency and computational efficiency need to be prioritized.

\section{Related Work}
\label{sec:related_work}

\subsection{Computational Humor Generation}
\label{sec:related_work:humor}
Computational humor is an interdisciplinary field at the intersection of NLP and humor theory. Early efforts in computational humor revolved around rule-based systems and linguistically-motivated methods. \newcite{Raskin1984SemanticMO} introduced a semantic analysis of humor which laid the foundation for subsequent rule-based approaches to humor recognition \cite{Mihalcea2005MakingCL, Reyes2012FromHR, chen-soo-2018-humor, weller-seppi-2019-humor}. Concerning the more challenging task of humor generation, early approaches were linguistically informed and focused on specific types of humor, \eg~puns or jokes \cite{ritchie-2005-computational, petrovic-matthews-2013-unsupervised}.

With the advent of deep learning, the focus shifted towards data-driven and neural approaches.  This ranges from using RNNs and GANs to create puns \cite{yu-etal-2018-neural, luo-etal-2019-pun},  to more recent transformer-based approaches \cite{garimella-etal-2020-judge} that can complete or generate jokes. These approaches enabled the generation of more contextually relevant and natural humor. Since neural approaches are primarily data-driven, this concurrently led to the creation of large joke datasets \cite{weller-etal-2020-humor} and benchmarks \cite{hossain-etal-2020-semeval}. 

Recent research has delved into the generation of figurative language such as sarcasm \cite{chakrabarty-etal-2022-flute}, puns \cite{mittal-etal-2022-ambipun}, as well as interactive chatbots with humor capabilities \cite{51348}. New multi-modal humor benchmarks \cite{hessel-etal-2023-androids} have been developed to gauge the humor understanding and explanation abilities of LLMs. \newcite{weller-etal-2020-humor}  proposed the first model for humor style transfer building a transformer model that translates from regular to humorous, and leveraging humor prediction data from news headlines for humor generation.

\subsection{LLMs as Evaluators}
\label{sec:related_work:eval}
Automatic evaluation of Natural Language Generation (NLG) tasks is typically based on N-gram metrics such as BLEU  \cite{papineni-etal-2002-bleu}, ROUGE \cite{lin-2004-rouge} and embedding-based metrics such as BERTScore \cite{bert-score} which require gold standard references.
Recent research has explored a reference-free approach to assess NLG tasks by leveraging the implicit knowledge and instruction following abilities of LLMs. \newcite{Fu2023GPTScoreEA} proposed GPTScore, which prompts LLMs with instructions and aspect definitions (\eg~fluency or coherence). The score is computed by calculating the conditional probability of generating the target text. GEval \cite{Liu2023GEvalNE} is an alternative metric, that presents the instructions in a form-filling paradigm and uses the probabilities of output tokens from LLMs to normalize the scores (\eg~score between 1 and 5) resulting in finer-grained continuous scores. In contrast to GPTScore and GEval, LLM-Eval \cite{lin-chen-2023-llm} uses a single prompt to evaluate multiple evaluation aspects, thus minimizing the calls to the LLM. Additionally, LLMs have also been used to assess the factuality of generated text \cite{zha-etal-2023-alignscore}. \newcite{mehri2023automatic} propose a learned evaluation metric via instruction tuning.
\subsection{Imitation Learning from LLMs}
\label{sec:related_work:imitation}
The emergence of LLMs has caused a paradigm shift in NLP from traditional knowledge distillation \cite{hinton2015distilling} to an imitation-based data or response distillation.  These approaches use LLMs as training data generators and train a smaller language model on this data \cite{west-etal-2022-symbolic}. With subsequent work demonstrating that this may be a sparse form of distillation, leading to the student mimicking the style of the teacher, but not the reasoning abilities  \cite{mukherjee2023orca, gudibande2023false}, further extensions have been developed to distill a complete ``Chain-of-Thought'' \cite{wei2023chainofthought} from the teacher model \cite{li-etal-2023-symbolic, wang-etal-2023-scott, shridhar-etal-2023-distilling, magister-etal-2023-teaching, hsieh-etal-2023-distilling}, improving the performance of smaller language models.
\subsection{LLM-based Alignment}
\label{sec:related_work:feedback}
A growing body of research leverages feedback from a language model to iteratively enhance its performance. This feedback may encompass written comments, numerical scoring, rankings, or explanations.  Humpback \cite{Li2023SelfAlignmentWI} aims to construct a better instruction-tuning dataset through an iterative self-training algorithm. 
Self-Refine \cite{Madaan2023SelfRefineIR} and REFINER \cite{paul2023refiner} explore LLMs to engage in self-reflection, providing feedback and encouraging the use of this feedback to enhance their responses. I2D2 \cite{bhagavatula-etal-2023-i2d2} demonstrates that small language models may improve if trained on their refined outputs generated using constrained decoding and filtered with a simple supervised critic model. 
Our work is inspired by BRIO \cite{liu-etal-2022-brio, liu2023learning} for summarization, which reuses the generation model as the evaluation model to rank the candidates with contrastive learning. However, we avoid a ranking-based approach and use pairwise feedback due to the challenges in mitigating positional and length biases among multiple generations. Concurrent to our work, Zephyr \cite{tunstall2023zephyr} uses preference learning to align a student model to user intent by using preferences derived from different candidate models from a large teacher.
 \begin{figure*}
    \centering
    \includegraphics[width=0.9\textwidth]{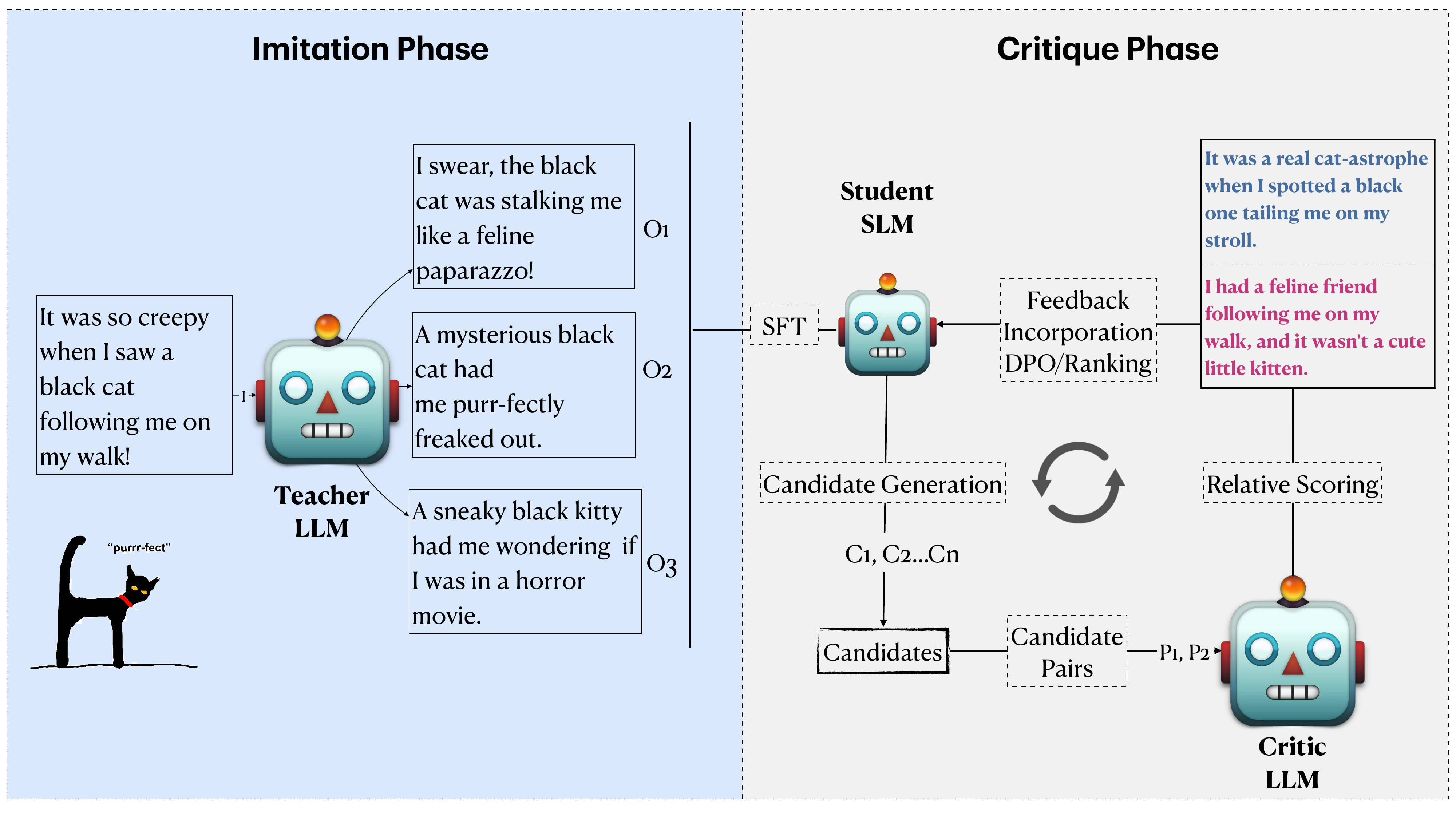}
    \caption{
The proposed knowledge distillation framework: We perform task-specific distillation from a large, general language model, in two phases: an initial imitation phase,  followed by a critical feedback phase which controls the quality of the generated humorous outputs from the student.}
    \label{fig:arch}
\end{figure*}



\section{Method}
\label{sec:method}
Our proposed knowledge distillation framework is depicted in Fig.~\ref{fig:arch}. Assuming that the teacher outperforms the student in the humor generation task, our objective is to bring the student's performance closer to that of the teacher.  We attempt to achieve this in two phases. First, in the imitation phase (Sec~\ref{sec:method:data}), the student undergoes finetuning using a set of humorous outputs $O$ = {$O_1, O_2, ..., O_n$} generated by the teacher for a given input $I$. In the critique phase (Sec~\ref{sec:method:arbiter}), the finetuned student generates candidate output pairs $P$ = {$P_1, P_2$}, which are then evaluated by a critic model. The scores obtained from the critic model are employed to train the student through a feedback-incorporation method. 
\subsection{Imitation Phase}
\label{sec:method:data}
During the imitation phase, we construct a dataset consisting of < literal, humorous > pairs and use it to directly train the student. As depicted in Fig.~\ref{fig:arch}, the process begins with a literal text input ($I$) and prompting the teacher to generate $N = 3$ humorous paraphrases ($O$) of the input in a zero-shot setting.
The utilization of $N$ outputs encourages diversity in student outputs, as it demonstrates multiple possible humorous paraphrases for a given input. The student is then finetuned on the constructed dataset by minimizing the cross entropy loss between the reference and predicted outputs. 
\subsection{Critique Phase}
\label{sec:method:arbiter}
In the critique phase, we aim to improve the student's task comprehension by teaching the student to differentiate between effective and ineffective humor. 
\paragraph{Critic}
The critic, an LLM (Large Language Model), evaluates the humor quality of outputs from the finetuned student.  Drawing from prior work on computational humor \cite{cd2c1a7919be46598073c41cac5ff742} and LLM-based scoring \cite{Liu2023GEvalNE, liu2023learning},  we adopt a pairwise relative scoring approach to mitigate subjectivity as opposed to providing an absolute score on a humorous output. To obtain the preferred output from the pair, we use Multiple Choice Prompting (MCP) \cite{robinson2023leveraging}, which, presents both paraphrases simultaneously and asks the LLM to pick the better humorous output. For two humorous paraphrases, P1 and P2, of the same input text I, we present them as candidates labeled with symbols (e.g., \quotes{1} and \quotes{2}). The critic LLM predicts a token ("1" or "2"), with associated probabilities indicating preference, which we denote as the Win Tie Rate (WTR) of P1 against P2. \footnote{The prompts used for WTR are shown in Appendix ~\ref{sec:appendix:prompts}.}

\paragraph{Feedback Incorporation}
\label{sec:method:feedback}
Subsequently, we leverage feedback from the pairwise scorer to refine the student model. Starting with the finetuned student (Fig.~\ref{fig:arch}), we generate a set of $k=6$ candidates using either diverse beam search or nucleus sampling. From this candidate pool, we select a pair of diverse humorous paraphrases, denoted as $P1$ and $P2$, where diversity is measured by a maximum pairwise n-gram-based edit distance score. The critic then scores these pairs, resulting in candidates categorized as either \textcolor{positive}{positive} or \textcolor{negative}{negative} based on their performance. Our objective here is to improve the finetuned student by discerning which output is preferred. To integrate this feedback, we explore the following feedback objectives.

\begin{enumerate}[leftmargin=*, itemindent=12px]
    \item  \emph{DPO} \cite{rafailov2023direct} provides an alternative to RLHF, aiming to align language models using human feedback without training an explicit reward model. When comparing two humorous paraphrases $P_i$ and $P_j$, if $P_i$ receives a higher quality score from the critic, DPO increases the likelihood of its completion over $P_j$. It employs the Bradley-Terry reward model, approximating reward modeling with a sigmoid loss function,

\begin{align*}
&\mathcal{L}_{\mathrm{DPO}}\left(\pi_\theta ; \pi_{\text{ref}}\right)=\\
&-\mathbb{E}_{\left(x, y_w, y_l\right) \sim \mathcal{D}}\left[\log \sigma\left(\beta \log \frac{\pi_\theta\left(y_w \mid x\right)}{\pi_{\text{ref}}\left(y_w \mid x\right)} - \right. \right. \\
&\left. \left. \beta \log \frac{\pi_\theta\left(y_l \mid x\right)}{\pi_{\text{ref}}\left(y_l \mid x\right)}\right)\right]
\end{align*}
where, $\pi_\theta$ represents the ratio of chosen to rejected scores from the fine-tuned LM, while $\pi_\text{ref}$ signifies the same ratio for an exact frozen copy of the model. The first term in the sigmoid denotes the shift in the preferred completion, while the second term indicates the shift in the dispreferred completion.
   

\item \emph{BRIO} \cite{liu-etal-2022-brio}, a sequence-level contrastive (ranking) objective proposed for abstractive summarization. For a given input $I$, when faced with two candidate humorous paraphrases, $P_i$ and $P_j$, and $P_i$ attains a higher quality score from the critic, the student is guided to assign a probability to $P_i$ that exceeds twice that of $P_j$.

\begin{math}
\begin{aligned}
\hat{\mathcal{L}}_{c t r}(\theta) & =\sum_{S_i, S_j \in \mathcal{S}_c, i<j} \max \left(0, \bar{p}_s\left(S_j \mid D ; \theta\right)\right. \\ 
& \left.-\bar{p}_s\left(S_i \mid D ; \theta\right)+\frac{1}{\lambda} \log 2(j-i)\right)
\end{aligned}
\end{math}

 where $\lambda$ corresponds to the average output length, and $\bar{p}_s$ corresponds to the length normalized log probabilities of paraphrases $i$ and $j$. Following \newcite{liu-etal-2022-brio},  we combine the cross-entropy loss used in the imitation phase with the margin loss as a multi-task loss, to maintain the generation abilities of the student.

\item \emph{BRIO-DPO}: We also experiment with combining both ranking and preference learning - for this variant, we obtain the contrastive pairs for the BRIO training from the teacher instead of the student, by letting the teacher rank two of its own responses. In this case, the BRIO loss is considered as supervision rather than feedback. The preference pairs for DPO still come from the student. To achieve this, we first train the student on BRIO, and use this model to initialize the DPO training.

We experiment with both one-shot and iterative feedback as shown in Figure ~\ref{fig:arch}, where the student may receive feedback from the teacher more than once.

\end{enumerate}

\section{Experimental Setup}
\label{sec:implementation}

 \paragraph{Dataset.}
Similar to FLUTE \cite{chakrabarty-etal-2022-flute}, we pick literal input sentences from the EmpatheticDialogues dataset \cite{rashkin-etal-2019-towards}.  Each conversation is obtained by pairing a speaker and a listener, where the speaker talks emotionally about personal matters, and the listener infers the underlying emotion and responds empathetically.  We sample 12,000 sentences from the training set and generate $N=3$ responses from the teacher to create 
36,000 literal-humorous text pairs. We sampled 1,000 sentences from the validation and test sets for evaluating the student and 100 samples from the test set for human evaluation. For out-of-distribution (OOD) evaluation, we sample 500 literal inputs from Samsum \cite{gliwa-etal-2019-samsum}.
\paragraph{Teacher Model.}
We use the 70B chat version of Llama 2 \cite{Touvron2023Llama2O} as the teacher model. We generate responses from the teacher using a temperature of 0.8 using nucleus sampling with top\_p = 1.

\paragraph{Student Model.}For student model,  we focus on BART-large \cite{lewis-etal-2020-bart}. All the BART student responses are obtained using beam search with the number of beams set to 5 during generation. 
\paragraph{Metrics.}
Traditional generation metrics such as ROUGE \cite{lin-2004-rouge}, may not work well for the creative task of generating humor. Inspired by prior work \cite{Liu2023GEvalNE, Fu2023GPTScoreEA}, we use the LLM-based metric of \textbf{Win Tie Rate (WTR)} also described in Sec \ref{sec:method:arbiter} for automatic evaluation. WTR aims to compare a pair of paraphrases and measures whether one paraphrase is equally (tie) or more (win) creative/humorous than the other, while preserving the meaning of the original input (consistency). In this work, we are more interested in humor than consistency - humor is essential, consistency can be subjective and optional based on the type of humor used (e.g. sarcasm). To make it more straightforward for human evaluation, we ask humans to provide individual WTR for the humor WTR separately from consistency WTR. We compute the automatic WTR of the student models using both the critic model based on Llama2-70B and GPT4~\cite{openai2023gpt4}.
\paragraph{Length and Positional Biases}
The critic is prone to two major biases - Position Bias, where evaluation changes based on the encoding
order of the responses and  Length Bias, the tendency to favor longer responses. 
To mitigate length bias, during the feedback training phase, we incorporate only a subset of the candidate training pairs,  filtered to maintain near-equal length between the two candidates (with a length ratio falling within the range of 0.8 to 1.2). This selection is intended to counteract the length bias resulting from the teacher's inclination towards longer outputs which has also been observed in prior work regarding AI \cite{Saha2023BranchSolveMergeIL} as well as human feedback\cite{Shen2023LooseLS}. To mitigate positional bias, we average all our win rates by letting the paraphrases take both position 1 and position 2.

\paragraph{Baselines.}

We finetune BART-large for 10 epochs on the teacher data using the Maximum Likelihood Estimation (MLE) objective (Sec.~\ref{sec:method:data}). This baseline is denoted as BART-FT-WS (WS stands for Warm Start). We further refine this baseline to create students trained without and with feedback losses.  For the first class of student models, we compare against,

\begin{itemize}
    \item \textbf{BART-FT}: BART-FT-WS further finetuned using the same MLE objective.
    \item \textbf{BART-SD}: Indicating Self-Distillation (SD), BART-FT-WS is trained on only the positive candidates obtained from the critic(Sec.~\ref{sec:method:feedback}).
\end{itemize}

For the proposed feedback-based baselines,

\begin{itemize}
    \item \textbf{BART-BRIO}: BART-FT-WS finetuned using the BRIO ranking objective for 10 epochs.
    \item \textbf{BART-DPO}: BART-FT-WS finetuned using the DPO objective for 10 epochs.
    \item \textbf{BART-BRIO-DPO}: BART-FT-WS finetuned using BRIO for 10 epochs, followed by DPO for an additional 10 epochs.
\end{itemize}




\section{Experiments}
\label{sec:results}
Our overall goal is to investigate whether a creative generation ability such as humor can be taught through a combination of examples from the teacher, and feedback on the student responses.  To answer this question, we focus our experiments on answering the following research questions:
\begin{enumerate}[label=\large\protect\textcircled{\small\arabic*}]
    \item\textbf{RQ1}: How does the teacher perform as a critic?
    \item \textbf{RQ2}: How do students trained with and without feedback compare?
    \item\textbf{RQ3}: How frequently should the teacher intervene?
    \item \textbf{RQ4}: How does data size affect student performance?
    \item \textbf{RQ5}: Does the length bias of the critic affect student responses?
\end{enumerate}

\subsection{\textbf{RQ1}: How does the teacher perform as a critic?}
In order to validate the role of the teacher as a critic model, we conduct blind human evaluation by some of the authors (``annotators''). In Table \ref{tab:he1}, we analyze the alignment between human and LLM evaluation and compare two versions of the critic. Cloze prompting scores each paraphrase separately based on the perplexity of the template ``The funny paraphrase of I is P$_i$'' (for $i = 1, 2$). Conversely, Multiple Choice Prompting (MCP) presents the two paraphrases to the model and instructs it to choose the better one. 


We ask the annotators to perform pairwise scoring of 100 blind pairs of humorous outputs. Annotators are asked to rank the paraphrases based on how humorous they are and how consistent they are with respect to the meaning of the input. We compare overall (WTR) and individual win tie rates for humor (WTR-H) and consistency (WTR-C) of the humorous outputs with the input using the critic model (Llama2-70B). We use the following metrics inspired by \newcite{saha2023branchsolvemerge} to evaluate the two scoring methods:
\paragraph{Ag\-H and Ag\-C.} We measure the  LLM-Human Agreement (Ag) which is a score between [0, 1] indicating the percent of pairs that the annotator and the LLM agreed on. We compute agreement by independently matching each human judgment for each pair with the model judgment.
\paragraph{Positional Bias (PB).} We measure the fraction of samples where the critic's scoring changes based on the order in which the paraphrases are presented to it.
\paragraph{Length Bias (LB).} LB aims to measure the model tendency to favor longer responses when the human did not.
 We measure length bias as the fraction of samples where humans prefer the shorter response, but the critic model prefers the longer response. 
 
 We observe that overall,  MCP performs significantly better in terms of human agreement when compared to the cloze style evaluation,  with an agreement on humor up to 76\% and consistency of up to 65\% across both forward and backward positions.  Concerning length bias, both methods exhibit length bias in about 20-25\% of the cases where they tend to prefer longer outputs than the humans. Although sub-metrics based on Humor (WTR-H) and Consistency (WTR-C) look promising, we found that combining them using simple (MEAN, AND) or more complex methods like BSM \cite{saha2023branchsolvemerge} results in amplifying positional biases to a huge extent for this task. Hence, we use the overall WTR as the automatic metric to evaluate model responses both during training and evaluation. 

\begin{table}[t!]
{%
\small
\begin{tabular}{cccclll}
\toprule
\textbf{Arbiter} & 
\textbf{Type} & 
\textbf{Ag-H$\uparrow$} & 
\textbf{Ag-C$\uparrow$} & 
\textbf{PB$\downarrow$} & 
\textbf{LB$\downarrow$}  \\
\midrule
Cloze & WTR-H & 57 & -  & 0 & 18  \\
Cloze & WTR-C & - & 46 & 0 & 21 \\
Cloze & WTR & 57 & 47 & 0 &  25 \\
MCP & WTR-H & 76 & - & 18 & 17  \\
MCP & WTR-C & - & 65  & 28 & 19  \\
\textbf{MCP} & \textbf{WTR} & \textbf{76} & \textbf{59}  & \textbf{15} & \textbf{20} \\
\bottomrule
\end{tabular}%
}
\caption{Evaluation of Human Agreement with different Arbiters - Agreement with Humor wins (Ag-H), Agreement with Consistency wins (Ag-C) and Positional Bias (PB), Length Bias (LB), Scores are multiplied by 100.}
\label{tab:he1}
\end{table}
 
\begin{table}[t!]
\centering
\small
{%
\begin{tabular}{llllrr}
\toprule
\textbf{Student} &   \textbf{WTR\textunderscript{GPT4}$\uparrow$}  & \textbf{WTR\textunderscript{llama2}$\uparrow$}   \\ \midrule
BART-FT    & 30 & 28  \\
BART-SD    & 35 & 36  \\ \midrule

BART-BRIO & 48& 53 \\
BART-DPO    & 52 & 60 \\
BART-BRIO-DPO  & 56 & 65 \\
\bottomrule
\end{tabular}

}
\caption{LLM-based Evaluation of student models  - Win Tie Rate(WTR) is measured against the teacher (Llama2-70B) by the critic llama2 (WR\textunderscript{llama2}) or external critic GPT-4 (WR\textunderscript{GPT4}) using the method described in Sec~\ref{sec:method:arbiter}. WTR is scaled by 100.}
\label{tab:main}
\end{table}
\begin{table}[t!]
\small
\centering
{%
\begin{tabular}{llllrr}
\toprule
\textbf{Student} &   \textbf{WTR\textunderscript{llama2}$\uparrow$}    \\ \midrule
BART-FT    & 34  \\

BART-BRIO & 42  \\
BART-DPO    & 47  \\
BART-BRIO-DPO  & 49  \\
\bottomrule
\end{tabular}

}
\caption{LLM-based Evaluation of student models on OOD test set (500 examples from \cite{gliwa-etal-2019-samsum}}
\label{tab:ood}
\end{table}
\begin{table}[t!]
\centering
\small
{%
\begin{tabular}{@{}lllr@{}}
\toprule
\textbf{Model} & \textbf{\# Data} & \textbf{Frequency} & \multicolumn{1}{l}{\textbf{WTR\textunderscript{llama2}}$\uparrow$} \\ \midrule
BART-FT & 36K & -            & 30 \\
BART-FT  & 24K & -            & 26 \\
BART-FT & 12K & -            & 24 \\ \midrule
BART-BRIO & 36K & 1/10 epochs  & 53 \\
BART-BRIO & 24K & 1/10 epochs  & 45 \\
BART-BRIO & 12K & 1/10 epochs  & 43 \\\midrule
BART-BRIO & 12K & 2/10 epochs  & 56 \\
BART-BRIO & 12K & 10/10 epochs & 66 \\ \bottomrule
\end{tabular}%
}
\caption{Effect of data size and frequency of feedback on win rate of student models against the teacher.}
\label{tab:analysis}
\end{table}
\newcommand{\dpo}{
\begin{tabular}{P{0.2\textwidth}P{0.25\textwidth}}
\toprule
\multicolumn{2}{l}{\begin{tabular}[c]{P{0.4\textwidth}} Input: I am a stay at home mom. \end{tabular}} \\ \midrule
\textbf{BART-FT} &\textbf{BART-DPO} \\ \midrule    
\textcolor{negative}{I'm a stay-at-home mom, and I'm loving every minute of it!} & \textcolor{positive}{I'm a stay-at-home superhero, saving the world one diaper change at a time. \vspace{0.7cm}                              }\\
\end{tabular}
}

\newcommand{\brio}{
\begin{tabular}{P{0.2\textwidth}P{0.25\textwidth}}
\toprule
\multicolumn{2}{l}{\begin{tabular}[c]{P{0.4\textwidth}} Input: i felt bad about sleeping in today. \end{tabular}} \\ \midrule
\textbf{BART-FT} &\textbf{BART-BRIO} \\ \midrule
\textcolor{negative}{I was feeling a bit guilty about my morning snooze-fest.} & \textcolor{positive}{I woke up feeling like a sloth who's been snoozed out for so long, I'm pretty sure I've heard the sound of snores echoing off the walls.}\\
\end{tabular}
}

\newcommand{\dpobrio}{
\begin{tabular}{P{0.2\textwidth}P{0.25\textwidth}}
\toprule
\multicolumn{2}{l}{\begin{tabular}[c]{P{0.4\textwidth}}Input: My son failed a really important test.\end{tabular}} \\ \midrule
\textbf{BART-FT} &\textbf{BART-BRIO-DPO} \\ \midrule
\textcolor{negative}{My son's test score was lower than a Kardashian's Instagram followers.} & \textcolor{positive}{My son's test results were so bad, I'm pretty sure he's been secretly practicing his 'I'm-not-a-complete-failure' face for weeks.}\\
\end{tabular}
}

\newcommand{\llamadpo}{
\begin{tabular}{P{0.2\textwidth}P{0.25\textwidth}}
\toprule
\multicolumn{2}{l}{\begin{tabular}[c]{P{0.4\textwidth}}Input: My wife and I are going to buy are very first brand new car this week\end{tabular}} \\ \midrule
\textbf{BART-DPO} &\textbf{BART-DPO-ITER2} \\ \midrule
\textcolor{negative}{We're finally upgrading our ride from a rusty old clunker to a sleek, sexy beast of a car.} & \textcolor{positive}{We're upgrading our transportation game from "legs" to "wheels" this week!}\\
\end{tabular}
}

\begin{table*}[t!]%
\small
  \centering
\subfloat[][]{\dpo}%
\subfloat[][]{\brio}\qquad
\subfloat[][]{\dpobrio}
\subfloat[][]{\llamadpo}
  \caption{Qualitative Examples showcasing variants trained with different types of feedback }%
  \label{fig:qe}%
\end{table*}
\label{sec:results:arbiter}
\subsection{\textbf{RQ2}:How do students trained with and without feedback compare?}
\label{sec:results:student}
Table~\ref{tab:main} shows the results of the automatic evaluation for the proposed student models using the Win Tie Rate (WTR) metric. To gauge the student's performance against the teacher that provided the training data, we 
compare the student outputs to the teacher's references. Hence, we measure the WTR of \emph{the student vs. the teacher}.  

When employing BRIO for ranking feedback, BART-BRIO surpasses the imitation-based students (BART-FT and BART-SD) with notable increases in performance—13-18\% and 19-25\% based on evaluations by both GPT-4 and Llama2-70B, respectively. This suggests that in approximately 50\% of cases, the student achieves comparable or superior performance to the teacher. When incorporating the same critic feedback as a preference learning objective through DPO (BART-DPO), we observe a similar trend of performance enhancement in the student, with the DPO baseline outperforming BRIO by a modest margin of 4-6\%. Subsequently, we explore a combined approach utilizing both BRIO and DPO objectives, resulting in a further improvement of 4-6\% over using DPO or BRIO independently.
\paragraph{OOD Test Set}
We show the performance of the BART-based student models on the OOD test set in Table \ref{tab:ood}. We can observe a similar trend of student performance improving with both BRIO and DPO feedback objectives. However, the BRIO-DPO variant results in a very minor performance boost indicating that further investigation is needed in combining ranking and preference learning objectives.
\subsection{\textbf{RQ3}: How frequently should the teacher intervene?}
\label{sec:experiments: frequency}
 To test whether the student can benefit from an increased frequency of feedback, we compare different versions in which we seek feedback from the teacher iteratively after every $K$ epoch which varies from 1, 5, and 10 (static). Table~\ref{tab:analysis}  shows that more frequent feedback has a positive impact on student performance, but the performance gains may saturate as feedback frequency is increased. There is also a trade-off between the communication cost with the teacher, which needs to be considered when providing more frequent feedback. This motivates future work in investigating this trade-off further, and approaches such as curriculum learning or active learning could be leveraged to choose which samples should receive feedback. 
\subsection{\textbf{RQ4}: How does data size affect student performance?}
\label{sec:experiments:datasize}
In this experiment, we assess the influence of the amount of data that is used to supervise the student.  We examine the relationship between data size and performance by varying data size between 4,000, 8,000, and 12,000 literal sentences, in all cases providing 3 humorous paraphrases per example.  Table~\ref{tab:analysis} shows that simply increasing the data size may be insufficient to improving the performance. The performance is lower than that of the more sophisticated teaching approach, specifically incorporating feedback, which proved to be more beneficial and data-efficient. We can observe that, the model with feedback trained on 12K examples still performed better than a model trained without feedback with triple the number of examples.
\subsection{\textbf{RQ5}: Does length bias of the critic affect student responses?}
The length bias in the teacher is propagated to the student, especially when trained over longer periods and more so with the BRIO or ranking objective.
Without any mitigation strategies, the length of the student outputs can almost double compared to the input text. As described in our Sec ~\ref{sec:method}, we also filter pairs that are significantly longer than one another during training to prevent the student from learning to optimize for length. This is similar to the length correlation observed in human feedback \cite{singhal2023long}. Addressing this issue requires addressing length bias in Multiple Choice Prompting in LLMs, which is an evolving research area \cite{Wang2023LargeLM, Shen2023LooseLS}.

\section{Qualitative Analysis}
\label{sec:analysis}
Figure~\ref{fig:qe} highlights instances where each of the feedback objectives (BART-DPO, BART-BRIO, and BART-BRIO-DPO) outperforms the BART-FT baseline trained solely through imitation learning. In Figure~\ref{fig:qe}a, BART-DPO produces a response that exhibits a higher degree of humor compared to BART-FT. In Figure~\ref{fig:qe}b, both BART-FT and BART-BRIO generate humorous responses, with BRIO's output slightly edging in humor but at the expense of being longer. Figure~\ref{fig:qe}c, BART-BRIO-DPO generates a more contextually relevant and amusing paraphrase in comparison to the BART-FT student. Lastly, Figure~\ref{fig:qe}d demonstrates an instance where the BART-DPO variant, receiving feedback for two iterations instead of one, exhibits superior performance compared to its counterpart trained with only one iteration of feedback.

\section{Conclusions}
\label{sec:conclusions}
We present a novel framework for knowledge distillation in the context of humor generation, with the teacher LLM providing references and feedback on a smaller student model's performance. Our approach involves leveraging a critic to guide the student model toward generating more humorous outputs. The effectiveness of our method is demonstrated through evaluations conducted by both the LLM and humans.  
 Additionally, we analyze the effect of various design choices on the performance, such as the frequency of feedback and training set size. 
 Our analysis sheds light on the limitations of LLM-based critics, and serves as motivation for future research in mitigation of biases in LLM-based evaluation and AI feedback.

\section{Limitations}
\label{sec:limitations}
\paragraph{Quantifying Humor} Measuring humor, especially in computational contexts, is inherently subjective and may not be accurately captured using simple metrics.

\paragraph{Cultural references} We observed that some of the generated humorous outputs were referring to celebrities and events in North America. As humor varies widely across individuals and cultures, the proposed models may not generalize well across diverse demographic and cultural groups.

\paragraph{Bias propagation} Training on feedback data from the LLM may lead to the propagation of any existing biases in the LLM’s training data. If the LLM’s data lacks diversity, these biases could be intensified in the smaller model. Similarly, the biases in model evaluation can also be propagated to the student.



\section{Ethical Considerations}
\label{sec:ethics}

\paragraph{Data.} The datasets used to gather the literal inputs outlined in Sec \ref{sec:implementation} are publicly accessible. Some of these datasets include crowdsourced annotations about emotional events, which may contain offensive, biased or hateful content. We use the chat variant of the llama2 that is aligned on generating safe warnings and responses when offensive content may be present. Further, the Llama2 model generates such warnings for about 3-4\% of our inputs, and we discared these inputs from the data distilled into the student.

\paragraph{Celebrity references.}  As shown in Figure ~\ref{fig:motivation}, the teacher LLM makes references and analogies to celebrities primarily from North America. When used as analogies for negative connotations, the humor generated may be considered offensive to specific people.


\bibliography{anthology,custom}
\bibliographystyle{acl_natbib}

\appendix
\label{sec:appendix}

\section{Pairwise breakdown of Human Scores}
\label{sec:appendix:he}
\begin{table}[ht]
\centering
\resizebox{\columnwidth}{!}{%
\begin{tabular}{@{}llllll@{}}
\toprule
\rowcolor[HTML]{FFFFFF} 
\textbf{Group} & \textbf{Baseline}  & \textbf{WTR-H $\uparrow$} &  \textbf{WTR-C $\uparrow$} & \textbf{WTR $\uparrow$} \\ \midrule
1 & BART              & 23 & 38  & 29 \\
2 & BART-BRIO         & 43 & 32  & 45  \\
3 & BART-DPO         & 51 & 34 & 50  \\
\end{tabular}%

}
\caption{Human Evaluation of student models against the teacher LLAMA2-70B. }
\label{tab:my-table}
\end{table}

\section{Prompts for Generation}
\begin{figure}[ht]
    \centering
    \small
    \includegraphics[width=\columnwidth, clip]{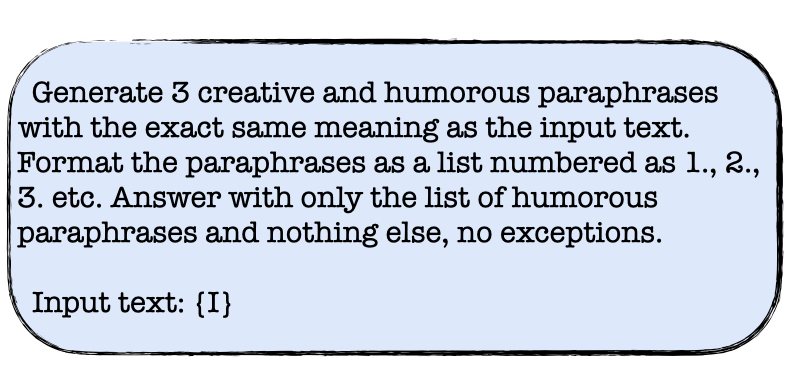}
    \caption{Humor Generation}
    \label{fig:prompt2}
\end{figure}

In Figure ~\ref{fig:prompt2}, we show the prompt used to obtain the humorous outputs from the teacher Llama2-70B model.
\label{sec:appendix:prompts}

\section{Prompts for Evaluation}
\begin{figure}[ht]
    \centering
    \small
    \includegraphics[width=\columnwidth, clip]{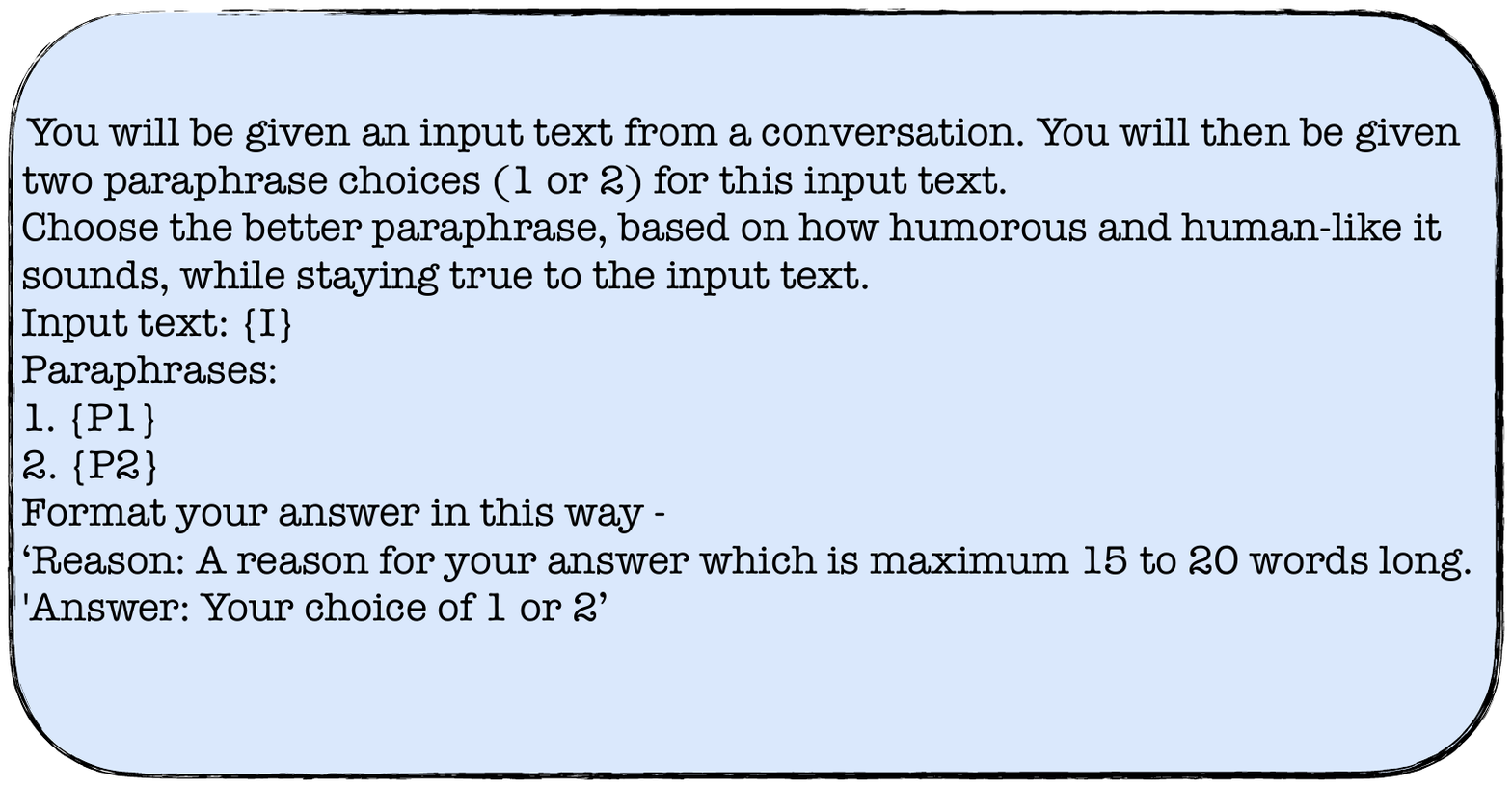}
    \caption{Pairwise evaluation using Multiple Choice Prompting}
    \label{fig:prompt1}
\end{figure}

In Figure ~\ref{fig:prompt1}, we show the prompt used to obtain the pairwise scores from the teacher Llama2-70B model.
\label{sec:appendix:prompts}



\end{document}